%% file: main.tex
\documentclass[10pt,twocolumn,letterpaper]{article}

\usepackage{multirow}

\usepackage{cvpr}              % To produce the CAMERA-READY version

\definecolor{cvprblue}{rgb}{0.21,0.49,0.74}
\usepackage[pagebackref,breaklinks,colorlinks,allcolors=cvprblue]{hyperref}

\def\paperID{*****} % *** Enter the Paper ID here
\def\confName{CVPR}
\def\confYear{2025}

\title{Entropy Bootstrapping for Weakly Supervised Nuclei Detection}

\author{James Willoughby$^{1}$\\
{\tt\small james.willoughby@cs.ox.ac.uk}\\
\and
Irina Voiculescu$^{1}$\\
{\tt\small irina.voiculescu@cs.ox.ac.uk}\\
\and 
$^{1}$University of Oxford\\
Department of Computer Science
}

\begin{document}
\maketitle
\input{sec/0_abstract}

\input{sec/1_intro}
\input{sec/1.5_related}
\input{sec/2_methods}
\input{sec/4_discussion}
\input{sec/5_conclusion}

\newpage
{
\small
\bibliographystyle{ieeenat_fullname}
\bibliography{main}
}

\input{sec/X_suppl}


\end{document}

%% file: sec/0_abstract.tex
\begin{abstract}
Microscopy structure segmentation, such as detecting
cells or nuclei, generally requires 
%mask annotations, where 
a human  to draw a ground truth contour around each
instance. Weakly supervised approaches (e.g.\ consisting of only single point labels) have the potential to reduce this workload significantly. 
Our approach uses individual point labels for an
entropy estimation to approximate an underlying
distribution of cell pixels. 
%It requires only a small percentage of the pixels that would be labeled in full masks. 
We infer full cell masks from this distribution,
and use Mask-RCNN to produce
an instance segmentation output. We compare this
point--annotated approach with training on the full
ground truth masks. We show that our method achieves a comparatively
good level of performance, despite a $95\%$ reduction in
pixel labels.
\end{abstract}

%We propose a pipeline for weakly supervised nucleus detection from point labels. This approach uses an entropy estimation step to approximate the underlying pixel distribution of the nuclei and is therefore based in robust uncertainty estimation. In addition this process is feasible using only $5\%$ of the pixels that would be labeled in full ground truth masks. We also provide a theoretical justification for how our process approximates the underlying pixel distribution. We then extend this estimated binary segmentation to full instance masks via a deterministic process and then train the instance segmentation network Mask-RCNN on these masks to produe a full instance segmentation output. We compare this with Mask-RCNN as trained on the full set of instanced masks to show that our method achieves a comparatively good level of performance with a $95\%$ reduction in pixel labels.

%% file: sec/1_intro.tex
\section{Introduction}
\label{sec:intro}

Histopathology is extremely important in detecting and assessing medical conditions, particularly in cancer~\cite{veta2014breast}\cite{humphrey2017histopathology}. As in many other areas of medicine Deep Learning has shown great potential to aid in computerised pathology, aiming to reduce both human error and human workload.

Cell instance segmentation has typically been the task of choice for deep learning based approaches, since the output provides the cell localisations and extent, which are easy for clinicians to verify and correct if needed. However the limitation behind such an approach is the vast amount of labeled data needed to train the models. These cell segmentation masks are especially time consuming due to the thousands of cells on any given slide, making expert annotation either prohibitively expensive or incredibly protracted. Even then, given the sheer number of cells we still expect these ground truths to be slightly inaccurate.

Due to the challenges involved in acquiring such a dataset, weakly supervised learning is a natural fit in this scenario, as it can significantly reduce the labeling requirement. We provide a novel approach for weakly supervised cell detection. It first uses uncertainty to ``bootstrap'' the weak labels, and then refines the ``bootstrapped'' predictions into individual cell detection instances.
% through a refining process.
%to turn the ``bootstrapped'' predictions into instance detections.

\begin{figure}
    \centering
    \includegraphics[width=\linewidth]{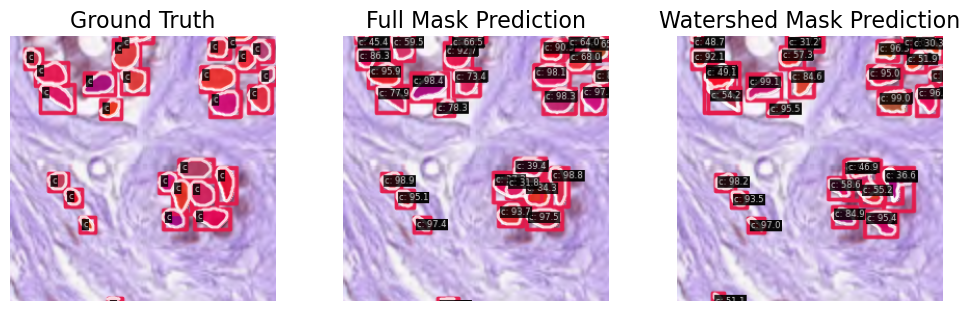}
    \includegraphics[width=\linewidth]{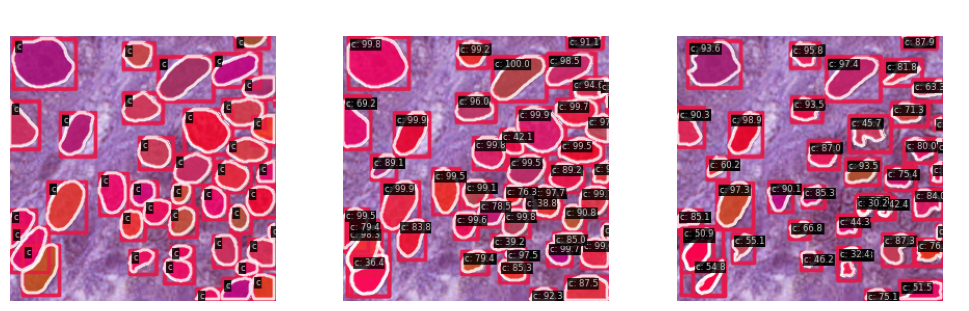}
    \includegraphics[width=\linewidth]{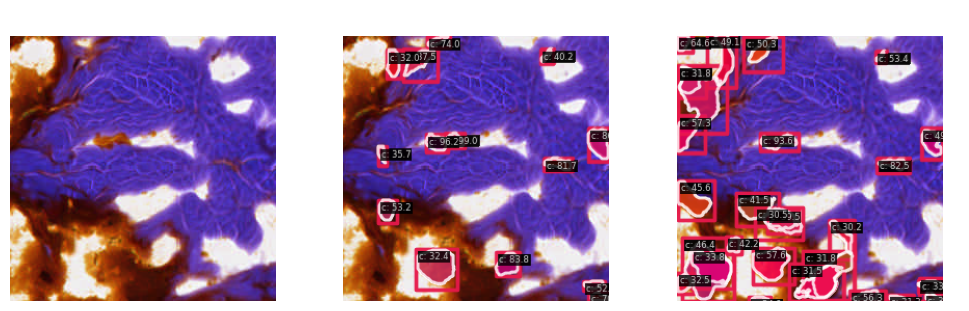}
    \caption{Mask-RCNN outputs. Left column: three separate images with ground truth masks overlaid. Middle column: predictions from training Mask-RCNN on full masks directly. Right column: predictions from training on our watershed instance masks as described in ~\cref{sec:maskrcnnassess}. The bottom image was chosen to show a universal failure case where both models detect false positives despite no nuclei being present.}
    \label{fig:step3assess}
\end{figure}

Rather than performing segmentation and checking mask overlap. we carry out {\em cell deection}.
%We motivate moving towards detection rather than segmentation as detection a
For clinical purposes, tasks such as predicting a bounding box for each cell
%high performing bounding box prediction 
provide a similar level of information, localising the cell within the image, giving an estimate of its size, and only omitting precise shape information. There is strong evidence that size analysis provides valuable information about pathology; by contrast, the prognostic value of shape analysis is in debate as reported in \cite{narasimha2013significance}. 

In addition to these clinical factors, there is also the issue of evaluation metrics. Many methods which take the segmentation approach report mean intersection over union (IoU) \cref{IoU} or the Dice score over the test set, which are useful for giving a general impression of segmentation overlap, but have drawbacks. Firstly the advantage of segmentation over detection would be to identify a precise boundary, yet Dice does not assess this well. Secondly since we care about segmenting individual instances, it is far more informative to use an instance-aware metric like segmentation mAP50, as described in ~\cref{sec:metrics step 2}. For these reasons as well as substantial differences in the datasets employed, comparisons against other work is not possible.  Ee are only able to compare our weakly supervised method with its fully supervised equivalent.

In addition we train and test our approach on a multi-tissue nuclei segmentation dataset which was designed to represent a more realistic clinical imaging scenario than other cell segmentation datasets often pose. 

%% file: sec/1.5_related.tex
\section{Related Work}

\subsection{Cell Detection and Segmentation}

The dataset and results we report are used for nuclei detection rather than full cell detection but, for obvious reasons, the methods involved in the two are closely related. As such we discuss methods relevant to both.

Automated Cell detection and segmentation methods have been an area of research for many years due to the large number of cells involved in histopathology as well as the subjectivity of the field. Watershed-based segmentation approaches have been historically popular such as those in \cite{gamarra2019split} and \cite{koyuncu2012smart} as these work well given a good contrast and marker proposal approach. Other popular approaches include Voronoi-based methods such as those in \cite{kaliman2016limits} and \cite{yu2010evolving}, since the Voronoi tiling of each cell's centroid is a good approximation of cell morpohology as discussed in more detail in \cite{kaliman2016limits}. The dual of the Voronoi diagram, the Delaunay triangulation, is also a useful classical procedure in this context and can be used as a way of separating clustered cells as described in \cite{wen2009delaunay}.

As Deep Learning has risen in popularity as a methodology, cell detection and segmentation has naturally adopted it as a strategy and has seen significant benefit. Supervised deep learning approaches have emerged that are able to achieve levels of accuracy similar to human annotators as described in \cite{greenwald2022whole} and \cite{sadanandan2017automated}. Both of these approaches address the issue of training data which is a conspicuous problem within this task and they do so in different ways. In \cite{sadanandan2017automated} the authors use an automatic training set generation process by converting fluorescent nuclei and cytoplasm into masks via the use of an existing tool. They also create a manually segmented test set but note that even this small proportion of their overall data is a significant number of cells to segment manually. In \cite{greenwald2022whole} they take an adjacent but distinct approach, gradually creating a dataset of one million labeled cells via a human-in-the-loop process. As these approaches show, there is considerable benefit to be gained from deep learning approaches to this problem but they inevitably run up against the issue of labeling.

\subsection{Weakly Supervised Cell Segmentation}

Given the obvious potential of deep learning approaches and the labeling limitation of traditional supervised learning approaches, many weakly supervised methods have been proposed to tackle this challenge.

Point or scribble annotations are the weak label of choice in the vast majority of methods as they are significantly easier to acquire than full masks and still provide a great deal of information about cell count and localisation which can be leveraged in order to acheive good results in cell detection and segmentation tasks.

Many methods approach the problem via a multitask approach. In \cite{chamanzar2020weakly} the network uses Repel encoding, clustering and a Voronoi diagram as combined losses for their network to optimise and \cite{qu2020weakly} uses a similar approach for their final loss function, optimising over cluster and Voronoi losses as well. This multitask approach allows for different aspects of the weak label information as well as any prior learned output to be leveraged efficiently.

Many approaches also base themselves around known information about cell morphology or distribution. In \cite{NucSeg_shapedict} for example the network is based around learning a set of shapes for cell boundaries in order to improve its precision around the cell border. In \cite{nishimura2019weaklydetresponse} the centriod of each cell is used as the training data which is then expanded upon in order to segment entire cells.

The Voronoi diagram of the cell annotations or predictions is used at some point in many methods as shown in \cite{qu2020weakly}\cite{chamanzar2020weakly}, because it provides a quick way to reduce the problem. This is because each point within a cell should generally be closer to its point label than any other point label. As a result only one instance will be contained within each partition of an accurate Voronoi diagram, which makes separating instances much easier.

Some methods approach the problem using a self-supervised or co-supervised training strategy as in \cite{zhao2020weakly} and \cite{tian2020weakly} this allows for a weak initial mask to be improved via consensus between simultanously trained networks.\\

\begin{figure*}[ht]
    \centering
    \includegraphics[scale=0.48]{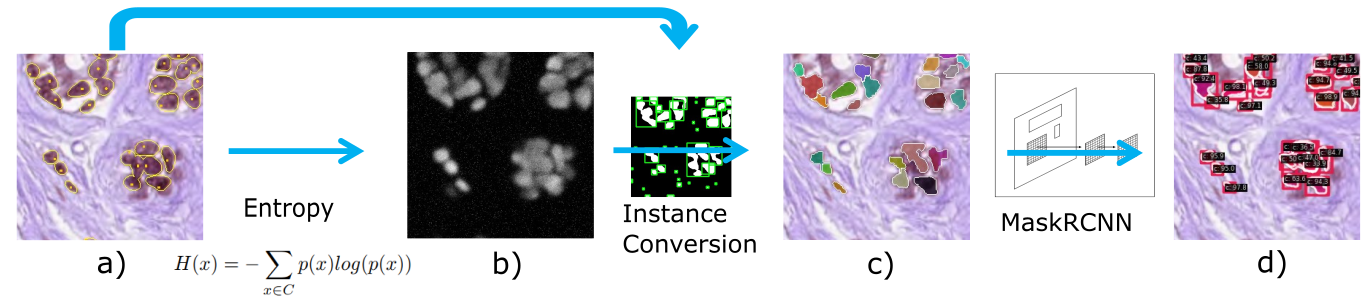}
    \caption{Figure showing the overall pipeline. The entropy estimation method we are using \cite{Carvalho_2020_CVPR} is trained on the point annotations shown in orange within a) to generate entropy shown in b). This is converted deterministically into instance masks as in c). Finally we use Mask-RCNN trained on these instance masks to detect nuclei as shown in d).}
    \label{fig:overallmethod}
\end{figure*}

\subsection{Our Contributions}

We distance ourselves from the above approaches in that we are primarily focusing on using a Bayesian Deep learning approach to ``bootstrap'' our point annotations to get to a high quality weak segmentation quickly. We then generate instance masks deterministically, and use these weak instances to train an off-the-shelf model (such as Mask-RCNN), to achieve our final detection outputs. Mask-RCNN is by no means the only option in this role, but we only report this one here because it is light--weight and it helps emphasise the overall principles.

While a probabilistic step does underpin a number of other methods, as seen in \cite{zhao2020weakly} and \cite{qu2020weakly} there is little discussion of the validity and quality of these initial processes. By using a robust Bayesian Deep Learning approach combined with random sampling of our input points, we generate an initial segmentation which correlates very well with the ground truth. We also provide a theoretical justification for this ``bootstrap'' approach to generating a weak segmentation, conveying the conditions under which this is valid.

%In addition to this our random sampling assumption is much weaker than assumptions made on the weak labels in other methods such as \cite{nishimura2019weaklydetresponse} in which the centroid of the cell is used as the weak label.

We go on to show that, using our weak segmentation, we can perform nuclei detection with good performance; this only uses a simple deterministic instancing process and conventional training with an off-the-shelf detection model. Our overall process is a lot simpler than something like a co-training infrastructure or learning a shape dictionary, and provides good detection results with only weak labels.\\

%% file: sec/2_methods.tex
\section{Methods}
\label{sec:methods}

\subsection{Data}
The data used in this paper was taken from the PanNuke dataset \cite{gamper2019pannuke} a cancer nucleus instance segmentation and classification dataset across $19$ tissue types. The central idea behind this dataset is that it aims to provide a dataset which represents the ``clinical wild'' better than existing histopathology datasets; methods trained on past datasets have tended to underperform when translated to real clinical scenarios. We believe that this kind of dataset represents the most challenging test for a weakly supervised approach. 

We used $2,656$ images from this dataset split $80:20$ into training and test sets. The images were all $256{\times}256$ pixels.

\begin{figure*}
    \centering
    \includegraphics[scale=0.39]{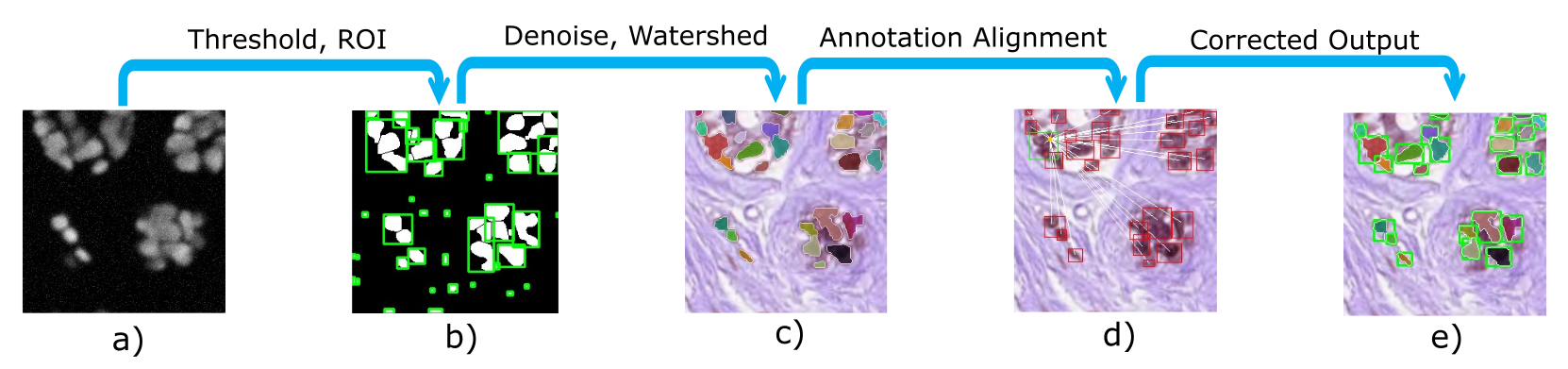}
    \caption{Figure showing the instance mask generation from the entropy shown in a). We use adaptive thresholding and the Voronoi edges to convert our distribution a) into a binary segmentation and we find the regions of interest (ROIs), the locations of the discrete connected elements of the segmentation. These are shown as green boxes in b). We then remove small artifacts and watershed to get instance masks as in c). These instance masks are then checked against the point annotations to remove erroneous instances as shown in d). Finally we output all the instances which are co-located with the point annotations, shown in e).}
    \label{fig:instanceseg}
\end{figure*}

\subsection{Entropy Bootstrapping}

The entropy bootstrap step is designed to take the point annotations provided and convert them into a ``prior'' segmentation which can then be crefined down into a more precise instance segmentation.

We first pass the images and point annotations into a Bayesian Segmentation Network in order to get a measure of the entropy over the labels. Specifically we used the Functional Variational Inference formulation outlined in \cite{Carvalho_2020_CVPR}. This network estimates the probability of each pixel belonging to each class $p(X=x)\forall x \in C$. In our case this is binary classification into the cases $x=\text{Nucleus}$ and $x=\text{Not Nucleus}$ for each pixel. This probability distribution over the classes of all the pixels which we can take the entropy given by:
\begin{equation}
    H(x) = -\sum_{x\in C}p(x)log(p(x)) \label{eq:Entropy}
\end{equation} 

This measures the average uncertainty in class prediction across the image.

Our point annotations are designed to replicate the scenario of an annotator looking over all the nuclei and placing a dot in each one. Given no guidance we assume that this dot could be placed in any random point within the nucleus.

In order to approximate this scenario with a clinician placing dots we take an existing dataset of instance segmented nuclei \cite{gamper2019pannuke} and randomly choose points within each nucleus to place our dots.  In practice we do this by sampling a random pixel within each nucleus and adding a small neighborhood around it to a binary segmentation mask. Given the random sampling of the pixel within the nucleus we assume that we can roughly express this ``sparsification'' of the ground truth labels by the conditional probability:
\begin{equation}
    \mathbb{P}(x_{L} = \text{Nucleus}|x_{T} = \text{Nucleus}) = \epsilon \label{eq:labelprob}
\end{equation}
This is the probability that a pixel will be labeled as Nucleus in our point annotations given that it is a Nucleus pixel in the ground truth masks. Here $x_{L}$ is the label of pixel $x$ in the sparse labels and $x_{T}$ is the label of pixel $x$ in the ground truth and $\epsilon$ is a small probability. Here ground truth can either crefer to a fully segmented set of masks or more abstractly to a clinician's understanding of the nucleus distribution. In the latter case this ``sparsification'' process is aiming to model the clinician placing a single dot in every nucleus they see in a set of slides.

We find empirically that when we use the labelling procedure, the images of the entropy of the model output closely resemble a binary segmentation of the nuclei against the background. Here we provide a brief theoretical justification for this. We also justify this with some qualitative examples and quantification in \cref{sec:step1assess}.

We make the assumption that the network we are using is a good estimator of the entropy over the input label distribution and we decompose this into:
\begin{equation}
 H = -\mathbb{P}(C_{L})log\mathbb{P}(C_{L}) - \mathbb{P}(C_{L}^{c})log\mathbb{P}(C_{L}^{c}) \label{eq:Entropy2}
\end{equation}
where $\mathbb{P}(C_{L})$ is the probability that pixel $x$ is labeled as \textit{Nucleus} and $\mathbb{P}(C_{L}^{c})$ is the probability that pixel $x$ is labeled as \textit{Background}. We can show that assuming \cref{eq:labelprob} is a good model for the labeling process used then for appropriately small $\epsilon$ the entropy in \cref{eq:Entropy2} behaves as:
\begin{equation}
    H(x) = \epsilon log(\epsilon) \mathbb{P}(C_{T})
\end{equation}
where $\mathbb{P}(C_{T})$ is the probability that a pixel $x$ has ground truth label \textit{Nucleus} and $\epsilon$ our small label probability. This derivation is outlined in Appendix A. This essentially allows us to use our good entropy approximator to produce a good binary segmentation of the nuclei with a minute fraction of the total pixel labels. Naturally there is a trade off as $\epsilon log (\epsilon)$ has to be small enough to be the dominant term in the entropy but large enough that the entropy estimation has enough samples to learn the distribution. 

In our method we sampled a single pixel at random in each mask and then added in a small neighborhood around it to mimic an annotator placing a dot with a stylus. We found that we got the best results with a neighborhood radius of 3. This corresponded to $\epsilon \approx 0.05$ meaning that we can get a reasonable approximation of $\mathbb{P}(C_{T})$ with a $95\%$ reduction in labeled pixels. See Appendix B for results.

\subsection{Instance Conversion}

The conversion process from our approximation of $\mathbb{P}(C_{T})$ to an instance segmentation has a number of issues to confront. Nuclei are often clustered and since we are usually looking at a two-dimensional slice through a three-dimensional tissue there can be overlapping nuclei. In addition to the separation issues from overlapping, we also have many cases where nuclei will be hard to distinguish from the background due to poor contrast. We may also have cases where the ground truth labels have missed some obvious nuclei amidst the large number of annotations.

In order to convert the distribution over the nucleus pixels into a set of instances we first apply a deterministic process, shown in \cref{fig:instanceseg}:

Firstly we apply gaussian blur over the entropy estimate to reduce the noise that will interfere with later steps. We then generate the Voronoi diagram of the input point annotations and take the edges. We subtract these edges from the entropy distribution in order to delineate between adjacent instances as much as possible.

We then apply gaussian adaptive thresholding to generate a binary segmentation of our separated entropy distribution, where the threshold value is given by a gaussian-weighted sum of values in the neighborhood of each pixel. This has to be done locally due to the nucleus clustering issue. The minima of our entropy distribution in regions with many nuclei are much higher than minima in more sparsely populated regions of the image. This results in a scenario where the contrast is different across the image, making adaptive thresholding required.

Given our binary segmentation we  generate a number of regions of interest across the image by finding the exterior contours of the segmentation. We take the bounding box around each contour to get the image regions which contain either discrete nuclei or close groupings of nuclei. Some contours will still contain clustered nuclei as the adaptive thresholding will not be able to separate everything.

We then apply a watershed process within each region of interest to further separate remaining close groupings of nuclei. The regions of interest are necessary since using a global distance transform approach can cause small nuclei to be eliminated when large clusters exist in the same image. We run the watershed fill over a gaussian smoothed version of the input image in order to generate masks that more closely resemble the smoother ground truth masks.

This set of masks  can still generate a large amount of unintended noise. The adaptive thresholding can result in small masks appearing in the background where there are local maxima of background noise. In addition to this there will be  some genuinely ambiguous masks that do not correspond to a label in the ground truth, but that could still be masking missed nuclei. The artifacts resulting from background noise are mostly removed via simple noise removal techniques such as erosion but rarely a few will still remain. In order to deal with these additional uncertain masks and the more stubborn artifacts we simply refer back to the initial nucleus labels and keep only the masks closest to each nucleus label, not allowing for double counting. This removes the uncertain labels which are not in the ground truth and any remaining artifacts. Since the objective of this process is to reproduce similar labels to the fully labeled masks removing the genuinely ambiguous masks is acceptable. They would have to be considered more carefully in the case of applying this network in practice.

\subsection{Training and evaluation using Mask-RCNN}
\label{sec:MaskRCNN}

Our instance masks generated by the watershed are likely to have rough borders and there is the potential of under- or over-segmentation of the nuclei in these masks which will interfere with alignment with the ground truth masks. As such we aim to use Mask-RCNN \cite{he2017mask} to try and improve these instance masks.

The rationale for why this should improve the masks is that if our watershed instances are good enough then they begin to look like ground truth masks which have been eroded by noise. From a learning perspective we expect that it should be easier for the network to learn the features that all the masks have in common, which represent good nucleus features, than it will be to learn the precise erosion needed to learn the watershed masks accurately. In particular Mask-RCNN has a multiscale feature embedding which allows for higher level features of the nucleus to be taken into account than the watershed is able to see.

As such we assume that given the watershed masks localise most of the nuclei and segment a significant proportion of each nucleus then we can use Mask-RCNN to improve the watershed masks significantly. This assumption of how Mask-RCNN will behave does rest on not introducing too many systematic errors across the watershed masks. This would introduce new easy-to-learn features which would harm performance as Mask-RCNN would learn to predict common mistakes. A good example of such an issue would be single masks over clustered nuclei, this kind of mistake would be easy to learn for Mask-RCNN and would compare very poorly against the ground truth.

One nice property of our work up until this step is that we have thus far used statistically grounded or deterministic processes giving the process a lot of transparency. We can propagate the entropy distribution all the way up to our watershed masks to give an uncertainty interpretation of our predictions however with the Mask-RCNN step we lose this. While Mask-RCNN does give a confidence output as part of its prediction, the confidence of a deep learning model is known to not be correlated well with uncertainty \cite{loftus2022uncertainty}. With a more uncertainty-based final step we could make our process more transparent and provide a reliable output uncertainty but this is reserved for future work.

\subsubsection{Implementation Details}
For our experiments we used Mask-RCNN \cite{he2017mask} with a number of different backbone choices, ResNet50 \cite{ResNet}, full size Swin Transformer \cite{swinT} and small size Swin Transformer. We used imagenet pretrained weights. We trained with the AdamW optimiser \cite{adamw} and linear learning rate decay with a base learning rate of $0.00008$.

%% file: sec/4_discussion.tex
\section{Results and Discussion}
\label{sec:discussion}
\begin{figure*}
    \centering
    \includegraphics[scale=0.68]{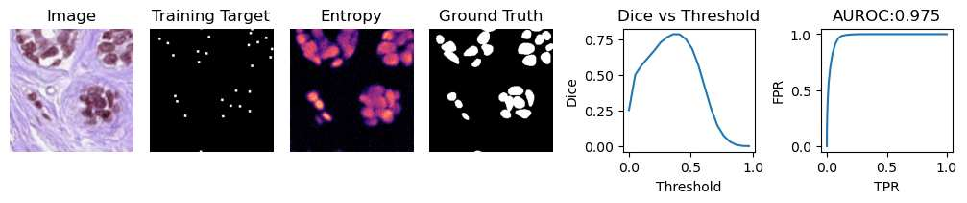}
    \includegraphics[scale=0.68]{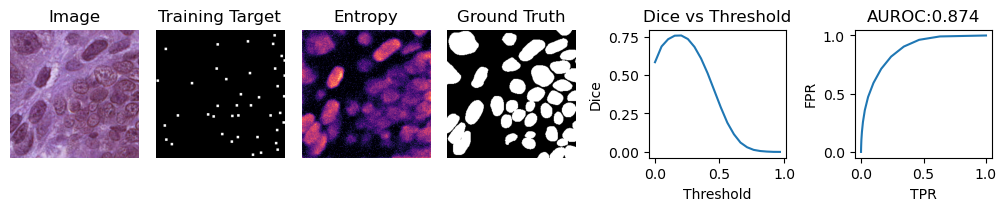}
    \caption{Figure illustrating assessment of entropy as an approximation of nucleus pixel distribution. We show here a test image, its corresponding point annotations, the test entropy and the ground truth. We also show how the Dice score varies with threshold value and an ROC plot.}
    \label{fig:step1assess}
\end{figure*}
\subsection{Assessing Entropy}

\textbf{Determining Appropriate Metrics}

In order to proceed with the process we have to assess the quality of our entropy distribution and in particular its suitability for use as a ``prior'' for our instance masks. As such we need to determine how well the distribution correlates with the ground truth masks.

We can threshold the entropy at different probability values in order to assess how well the distribution matches the ground truth at these different thresholds. A straightforward way to do this is to take the Dice score of the thresholded entropy when compared with the background. Plotting these Dice values against the entropy threshold we can produce a graph as in the left plot in \cref{fig:step1assess}. The limitation of this Dice measure when we are mostly interested in detection and localisation is that our Dice score will be penalised at higher thresholds as we will undersegment each nucleus however we will still be accurately localising the nucleus. 

In order to assess how well the entropy discriminates between nucleus and background on a distribution level we are motivated to also include an AUROC like measure, shown as the rightmost plot in \ref{fig:step1assess}. The Dice measure has limitations as discussed, however the AUROC measure will not penalise the method for undersegmentation in the same way. This measure also has its flaws however as it is biased due to the large number of background pixels which are easy to classify. This will skew the numbers higher provided the entropy classifies the background well.\\

\begin{figure*}
    \centering
    \includegraphics[scale=0.33]{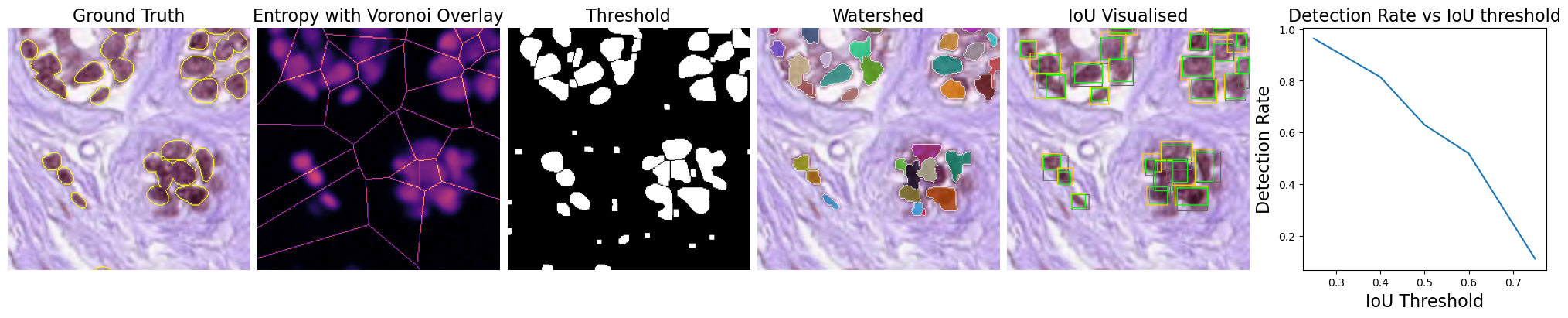}
    \includegraphics[scale=0.33]{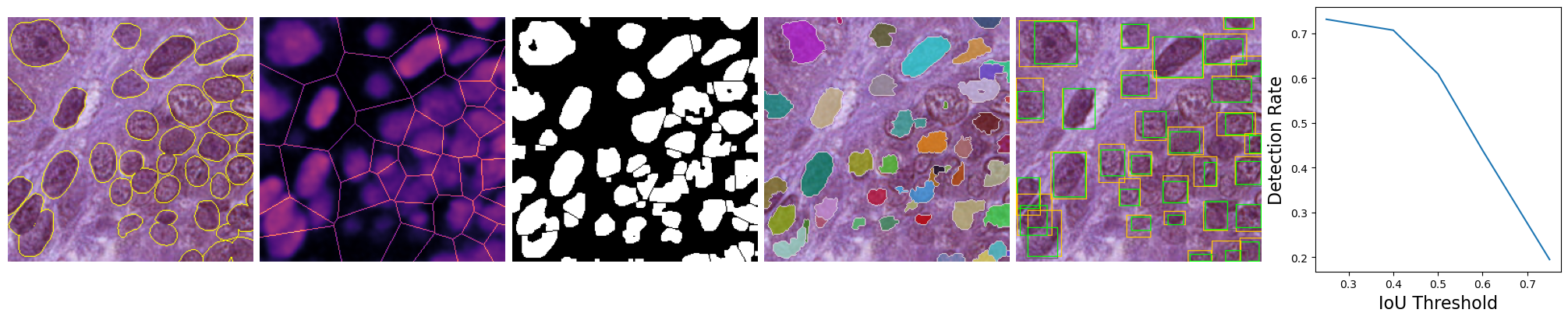}
    \caption{Figure illustrating assessment of instance masks generated from entropy. We show a test image with the ground truth instance masks outlined in orange, the test entropy overlaid with the Voronoi edges, the binary mask from adaptive thresholding and the output of the watershed instancing. We then show a visualisation comparing the ground truth bounding boxes in orange with the watershed bounding boxes in grey. The intersection of the two is shown in green. Finally we show a plot of nuclei detection rate against IoU threshold.}
    \label{fig:step2assess}
\end{figure*}

\textbf{Results of these Metrics}
\label{sec:step1assess}

The Dice plot is useful for comparing how well the entropy matches the ground truth in the best case scenario by taking the peak. It is also a good sense check that the entropy and the ground truth correlate strongly. We plot these graphs for all test images and take a mean across the entire test set and get a peak Dice score of $0.71$, indicating that the entropy correlates well across the entire test set.

For the AUROC measure we find that the entropy distribution produces very high AUROC scores indicating good separation of the nucleus and background classes. We would expect this given our theoretical motivation but these results back this up quantitatively. Across the entire dataset we get a mean AUROC of $0.944$, which we must be aware is a biased measure but is still strong evidence that the method discriminates between nucleus and background well across the entire dataset.

Qualitatively the AUROC is also a better assessor of the entropy's performance than the Dice score. We can see in the second row of \cref{fig:step1assess} that the entropy distribution has uniformly higher values in the area of the image with poorer contrast and the nuclei are marked out much less distinctively. This poor qualitative performance is creflected in the poorer AUROC but not so clearly in the Dice plot.

We can also use this analysis to compare the effect of different sampling approaches and artificial errors. For example we can choose to omit a certain percentage of point annotations or we can choose to add noise to the positions of the points, further weakening the labels. Reducing our number of labels by half only reduces the AUROC to $0.9$, indicating that the entropy still functions as a good discriminator at half labels even when taking into account the bias in this measure. However the mean Dice peak decreases to $0.6$ indicating a significant loss of precision as would be expected when reducing the labels this much. See Appendix B for full results. 

Introducing a random pixel position error into our labels of up to 5 pixels in any direction reduced our AUROC to $0.926$ and our mean Dice peak to $0.64$ once again indicating the loss of precision we would expect and that the entropy is still a good discriminator. This assessment of behaviour under these types of errors is useful as it simulates the main type of error we might expect when sourcing a large number of point annotations of nuclei, missed instances and incorrect placement of points. As such it is valuable that our entropy approach remains well correlated with the ground truth even under a significant amount of error.

\subsection{Assessing Deterministic Instance Generation}
\textbf{Determining Appropriate Metrics}
\label{sec:metrics step 2}

In order to assess how well our instance generation step performs at generating instance masks we need to compare with the ground truth instance masks. A natural way to do this is via an mAP50-like metric since this is a widely used standard for object detection, however we lack confidence values for our instance masks so direct use of the metric is not possible. We instead choose to measure the detection rate at different IoU thresholds across the test set. IoU is the intersection over union, measuring the extent to which two regions match as:
\begin{equation}
    IoU = \frac{x_{pred} \cap x_{gt}}{x_{pred}+x_{gt}-x_{pred} \cap x_{gt}}\label{IoU}
\end{equation}

Here $x_{pred}$ and $x_{gt}$ are the sets of predicted pixels and ground truth pixels respectively. Detection rate is given by:

\begin{equation}
    D_{\alpha} = \frac{TP_{\alpha}}{N_{nuclei}}\label{eq:detrate}
\end{equation}

Where $TP_{\alpha}$ is the number of predictions which overlap with a ground truth box with IoU greater than threshold value $\alpha$ and $N_{nuclei}$ is the total number of nuclei in the ground truth.

We calculate this value for a number of different thresholds from $0.25$ to $0.75$. We expect to perform poorly near $0.75$ as this indicates precise alignment of the predictions and the ground truth which we do not expect to achieve for a number of reasons. Firstly we are using watershed on the image pixels to generate our masks which will result in jagged boundaries which will never align precisely with the smooth boundaries given in the ground truth. Secondly, even though we are confident that we are separating the nucleus clusters well, given our Voronoi diagram is generated from random points within the nuclei we are unlikely to be separating exactly on the nucleus boundary, causing imprecise bounding boxes.

We expect to perform well at a threshold value of $0.25$ since this is the most forgiving to imprecise prediction boxes however this is a poor metric of detection since it will allow for a prediction of a whole cluster of nuclei to be marked as correct in many cases and we do not want to reward this behaviour. As such a prediction threshold of $0.5$ is considered to be a reasonable metric to base assessment around. This threshold indicates that a prediction is well localised and scaled compared to the nucleus and minimises correct predictions for nucleus clusters but allows for detection with the imprecise nucleus borders we expect to see. This justification is backed up by mAP50 being considered one of the core assessment metrics for object detection in general \cite{everingham2010pascal}.\\

\textbf{Results of these Metrics}
We show some qualitative examples of the outputs in \cref{fig:step2assess}. We can see from these outputs that we perform poorly at a threshold value of $0.75$ as predicted. From the qualitative outputs and the significant increase in performance if we take the threshold at $0.5$ we can see that this is not due to poor localisation performance but due to the lack of precision we predicted. From the qualitative results, particularly the second row of \cref{fig:step2assess} we can see that the nucleus is undersegmented in many cases, which leads to missed detections for larger nuclei. 

When we take the average detection rates over the entire set of images we find that at a threshold of $0.5$ we get a mean detection rate of $0.61$ indicating that we are detecting the nuclei at a good rate across the entire set of images by just applying a corrected watershed to our entropy distribution. We anticipate that using an instance detection method we could improve this however, as outlined in \cref{sec:MaskRCNN}

\subsection{Assessing Mask-RCNN output} 
\label{sec:maskrcnnassess}
\textbf{Determining Appropriate Metrics}

Mask-RCNN outputs are generally assessed on an instance basis using the bounding box and segmentation mAP measures. For bounding box the IoU is calculated by comparing the masks and for segmentation it is done by comparing the masks directly. Naturally we expect the bounding box performance to be better than the segmentation performance and since fundamentally we are interested in detection and localisation we can treat the bounding box metric as our main metric of quality. As discussed in \cref{sec:metrics step 2} we are interested mostly in the values at IoU greater than $0.5$ so we report the values for thresholds of $0.5$ and $0.75$ as shown in \cref{tab:1} and \cref{tab:2}.\\

\begin{table}[]
\begin{tabular}{lllll}
\multirow{2}{*}{Labels} & \multirow{2}{*}{Model} & \multicolumn{3}{l}{Bounding Box} \\ 
                        &                        & mAP50           & mAP75  & mAP   \\\hline\hline
\multirow{3}{*}{GT (100\%)}    & Resnet50               & 0.800             & 0.552  & 0.498 \\
                        & Swin-S                 & 0.838           & \textbf{0.604}  & \textbf{0.537} \\
                        & Swin-T                 & \textbf{0.834}  & 0.592  & 0.529 \\ \hline
\multirow{3}{*}{Weak (5\%)}   & Resnet50               & 0.715           & \textbf{0.300}    & 0.349 \\
                        & Swin-S                 & 0.719           & 0.295  & 0.347 \\
                        & Swin-T                 & \textbf{0.724}  & 0.299  & \textbf{0.351}
\end{tabular}
\caption{Bounding box ablation study of backbones for Mask-RCNN. Best performance for each label type, Ground Truth (GT) and Weak, bolded.}
\label{tab:1}
\end{table}

\begin{table}[]
\begin{tabular}{lllll}
\multirow{2}{*}{Labels} & \multirow{2}{*}{Model} & \multicolumn{3}{c}{Segmentation}                 \\
                        &                        & mAP50          & mAP75          & mAP            \\ \hline\hline
\multirow{3}{*}{GT (100\%)}    & Resnet50               & 0.783          & 0.499          & 0.462          \\
                        & Swin-S                 & \textbf{0.828}          & \textbf{0.547}          & \textbf{0.499}          \\
                        & Swin-T                 & 0.826 & 0.544 & 0.496 \\ \hline
\multirow{3}{*}{Weak (5\%)}   & Resnet50               & 0.629          & \textbf{0.164}          & 0.262          \\
                        & Swin-S                 & 0.628          & 0.163          & \textbf{0.259}          \\
                        & Swin-T                 & \textbf{0.632} & 0.157 & 0.258
\end{tabular}
\caption{Segmentation ablation study of backbones for Mask-RCNN. Best performance for each label, Ground Truth (GT) and Weak, type bolded.}
\label{tab:2}
\end{table}

\textbf{Results of these Metrics}

We can see in \cref{fig:step3assess} a qualitative comparison of training on our watershed masks versus training on the ground truth masks. We note from the images that our method seems to localise the nuclei at a similar level to training on the full masks and this is backed up by our mAP50 performance of $0.724$ compared to the full mask performance of $0.834$, indicating that we are indeed localising to a similar extent across the entire test set. When we look at the mAP75 performance we see a drop off in performance, with our method achieving a score of $0.3$ compared to the full mask score of $0.604$. This is to be expected for two reasons. For one we are not training with the ground truth boxes as a target so it is not reasonable to expect precise alignment with them. In addition we note that even the approach trained on the full ground truth only achieves a score of $0.604$, likely a manifestation of uncertainty in the data which we know to be there due to the inaccuracies in the ground truth.

As such we focus mainly on the good localisation performance seen with the mAP50. We note from \cref{fig:step3assess} that some instances are missed in our method and some nuclei are undersegmented which could be contributing to the gap in performance. The performance of Mask-RCNN overall could definitely be improved also, in the third row of \cref{fig:step3assess} we see both methods predict instances where there are none in the ground truth. This kind of false positive prediction is clearly part of the behaviour of Mask-RCNN in this task and these errors need to be eliminated.

%% file: sec/5_conclusion.tex
\section{Conclusions and Further Work}
\label{sec:conclusion}
In addition to proposing a novel weakly supervised nuclei instance detection method based on entropy estimation, we also wish to generate discussion over the appropriate choice of metrics for assessing such tasks. 

We show that the masks generated with our relatively simple method can be used with an off-the-shelf instance segmentation model to approach the performance seen when training with the full set of original instance masks. We also find that Mask-RCNN has limitations in this scenario, indicating that better results could be achieved with a domain--specific final step, to be investigated in future.

In addition we would like to investigate a more uncertainty-focused approach which can take advantage of the entropy distribution better or provide grounded uncertainty estimates of output predictions.

Another advantage of an uncertainty based approach would be the ability to deal with ambiguous or missed labels in a more principled manner. This would enable an approach better suited to the reality of nuclei detection.

We can also extend this to multiclass detection fairly easily if our point annotations have class labels attached as we can reintroduce this information when making the watershed masks. This would provide valuable additional information in the output.

%% file: sec/X_suppl.tex
\clearpage
\setcounter{page}{1}
\maketitlesupplementary

\section{Appendix A}
\label{sec:Appendix A}

Starting from our entropy definition of:
\begin{equation}
    H = -\sum_{x\in C}p(x)log(p(x)) \label{eq:1}
\end{equation}
We use the fact that we are doing a binary labelling process to define:
\begin{align}
\mathbb{P}(x_{\text{L}}=\text{nucleus}) &= \mathbb{P}(C_{L})\\
\mathbb{P}(x_{\text{L}}=\text{background}) &= \mathbb{P}(C_{L}^{c}) = 1-\mathbb{P}(C_{L})\\
\mathbb{P}(x_{\text{T}}=\text{nucleus}) &= \mathbb{P}(C_{T})\\
\mathbb{P}(x_{\text{T}}=\text{background}) &= \mathbb{P}(C_{T}^{c}) = 1-\mathbb{P}(C_{T})    \label{eq:probdef}
\end{align}

Where $x_{L}$ is the label of pixel $x$ which is input into the network and $x_{T}$ is the ground truth label of pixel $x$.

Our labeling approach is such that we accurately label all the background pixels. As such we make the assumption that given a background pixel from the ground truth, we label it correctly in the input labels with probability $1$:
\begin{equation}
\mathbb{P}(C_{L}^{c}|C_{T}^{c}) = 1  \label{eq:2}
\end{equation}
The second component of our approach is that we label the foreground or nuclei pixels accurately only with small probability $\epsilon << 1$. So our probability that a nucleus pixel in the ground truth will be labeled as a nucleus pixel in our final labels is:
\begin{equation}
\mathbb{P}(C_{L}|C_{T}) = \epsilon    \label{eq:3}
\end{equation}
Expanding the entropy we get:
\begin{equation}
    H = -\mathbb{P}(C_{L})log\mathbb{P}(C_{L}) - \mathbb{P}(C_{L}^{c})log\mathbb{P}(C_{L}^{c})  \label{eq:4}
\end{equation}

We can use Bayes theorem to rewrite $\mathbb{P}(C_{L})$ as:
\begin{equation}
\mathbb{P}(C_{L}) = \mathbb{P}(C_{L}|C_{T})\mathbb{P}(C_{T}) + \mathbb{P}(C_{L}|C_{T}^{c})\mathbb{P}(C_{T}^{c})\label{eq:PCL}
\end{equation}
and we can use that $\mathbb{P}(C_{L}|C_{T}^{c})$ is the complement of our assumption in \ref{eq:2} to show:
\begin{equation}
\mathbb{P}(C_{L}|C_{T}^{c}) = 1 - \mathbb{P}(C_{L}^{c}|C_{T}^{c}) = 0 \label{eq:PCL|CLC}
\end{equation}

which we substitute into \ref{eq:PCL} get:
\begin{align}
    \mathbb{P}(C_{L}) = \epsilon\mathbb{P}(C_{T})    \label{eq:5}
\end{align}

This now allows us to rewrite our entropy from \ref{eq:4} as:
\begin{equation}
    H = -\epsilon\mathbb{P}(C_{T})log\epsilon\mathbb{P}(C_{T}) - (1-\epsilon\mathbb{P}(C_{T}))log(1-\epsilon\mathbb{P}(C_{T}))
\end{equation}
substituting $x = \mathbb{P}(C_{T})$ and rearranging we can then write the entropy in terms of $\epsilon$ and $x$:
\begin{equation}
    H = -\epsilon x log\epsilon - \epsilon x log x - (1-\epsilon x)log(1-\epsilon x)
\end{equation}
In order to see which term dominates in this expression for small $\epsilon$ we look at the relative rates at which each term tends to zero as $\epsilon \rightarrow 0$. We can see that $\epsilon x log(x)$ goes to zero faster than $\epsilon x log(\epsilon)$ as:
\begin{equation}
lim_{\epsilon \rightarrow 0}\frac{\epsilon x log\epsilon}{\epsilon x log x} = lim_{\epsilon \rightarrow 0}\frac{log\epsilon}{log x} = 0 \label{eq:lim1}    
\end{equation}

and similarly we can show that:
\begin{equation}
lim_{\epsilon \rightarrow 0}\frac{(1 - \epsilon x )log(1-\epsilon x)}{\epsilon x log x} = 0 \label{eq:lim2}    
\end{equation}

This means that for sufficiently small $\epsilon$ the $-\epsilon x log \epsilon$ term dominates and the entropy behaves as:
\begin{equation}
H = -\epsilon log(\epsilon) \mathbb{P}(C_{T})
\label{eq:Hpenul}    
\end{equation}

For small $\epsilon$ this $\epsilon log (\epsilon)$ term is a small positive constant, so we can then write:
\begin{equation}
H \propto \mathbb{P}(C_{T}) \label{eq:Hfinal}    
\end{equation}

As observed.

\newpage

\section{Appendix B}
\label{sec:Appendix B}

We also report the results of ablation studies into varying parameters in the sampling procedure. 

\subsection{Varying Point Radius}
In our point annotation creation procedure we sample our random point from within each ground truth instance mask and then convert it into an input mask by adding a neighborhood around the point. This is done to mimic point labels that a clinician would create with a stylus when labeling nuclei. Our choice of radius for this virtual ``pen'' has two effects. If the radius is too large then it will likely create a mask which spills over the nucleus border, introducing additional uncertainty, but also introducing more total correct pixels into the labels. If the radius is too small then the network may have too little information to learn on, resulting in a poor output.

\begin{table}[!h]
\begin{tabular}{lll}
Radius (Pixels) & Mean Peak Dice & Mean AUROC \\\hline\hline
1               & 0.37           & 0.691      \\
3               & 0.71           & 0.944      \\
6               & 0.683          & 0.942      \\
10              & 0.651          & 0.932     
\end{tabular}
\caption{Results of point annotation radius tests on Peak Dice and AUROC.}\label{tab:rad}
\end{table}

\begin{figure*}[!h]
    \centering
    \includegraphics[scale=0.6]{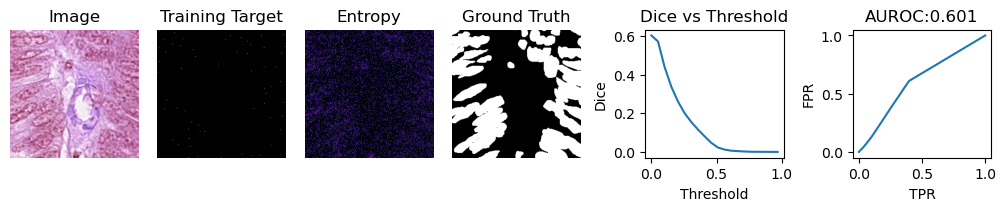}
    \includegraphics[scale=0.6]{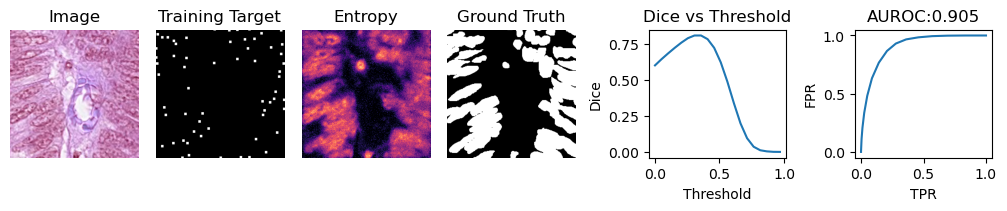}
    \includegraphics[scale=0.6]{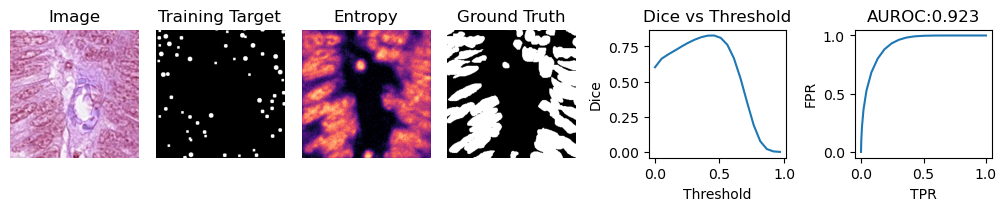}
    \includegraphics[scale=0.6]{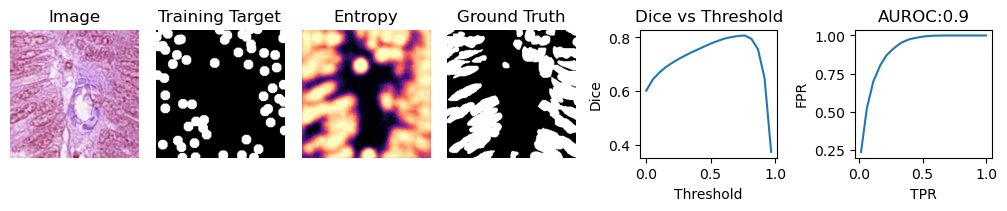}
    \caption{The effect of varying point annotation radius on entropy output. The first row shows $r=1$, second shows $r=3$, third shows $r=6$ and fourth shows $r=10$.}
    \label{fig:AppBexprad}
\end{figure*}

We tested our point at radii varying from $1$, just the point on its own, up to $10$ pixels. Visuals of these experiments are shown in \cref{fig:AppBexprad} and results shown in \cref{tab:rad}. We find that a radius of $3$ pixels gave the best results. As shown in the figure, a radius of $1$ pixel seems to be too little data for the network to learn a good distribution and a radius of $10$ pixels produces an oversegmentation of the nuclei. A radius of $6$ pixels performs similarly but slightly worse than $3$ pixels. So we choose $3$ pixels as our point annotation radius both as it gives the best performance and has the best trade-of between number of pixels annotated and performance.

\subsection{Percentage of Point Annotations}
We also carried out a study into the effect of reducing the percentage of point annotations used in training on entropy performance. This represents another axis along which labels can be made ``weaker'', by reducing the number of instances labeled at all. We expect that decreasing the percentage of instances labeled will decrease the efficacy of our method but it is interesting to see how the entropy varies as this is changed. Of course because there are a very large number of nuclei labeled across the entire dataset, well over $100,000$, a reduction by $50\%$ still leaves us with a very large number of labeled nuclei.

\begin{table}[]
\begin{tabular}{lll}
Annotations   (\%) & Mean Peak Dice & Mean AUROC \\\hline\hline
100                      & 0.71           & 0.944      \\
70                       & 0.64           & 0.925      \\
50                       & 0.62           & 0.904     
\end{tabular}
\caption{Results of varying the percentage of point annotations used in training on Peak Dice and AUROC.}\label{tab:sample}
\end{table}

\begin{figure*}[!h]
    \centering
    \includegraphics[scale=0.6]{images/Appendix_B/R=3/uncertainty_1516.png.png}
    \includegraphics[scale=0.6]{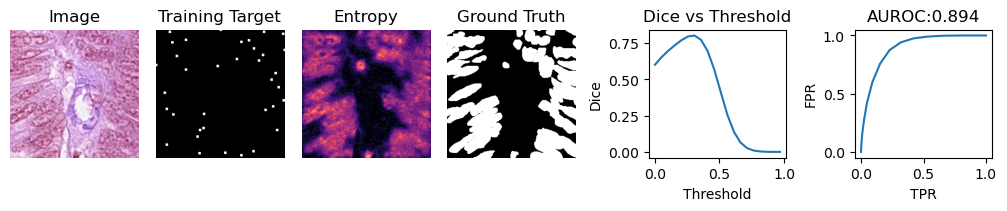}
    \includegraphics[scale=0.6]{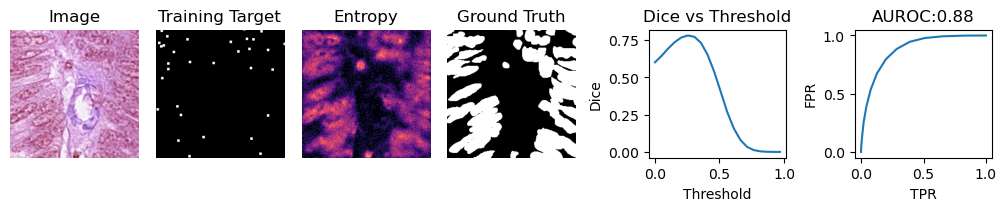}
    \caption{The effect of varying the percentage of point annotations trained with on entropy. First row shows $100\%$, second shows $70\%$ and third shows $50\%$.}
    \label{fig:AppBexpsample}
\end{figure*}

Visual outputs of this experiment are shown in \cref{fig:AppBexpsample} and numerical results are shown in \cref{tab:sample}. We see that varying the percentage of labeled nuclei down to $50\%$ does not actually change the performance as drastically as we might expect. This is presumably due to the large number of nuclei in the database as mentioned, tens of thousands of nuclei is still a significant number for the network to learn from. We chose to proceed using all the nucleus labels as this gave the best performance and also made comparison with the ground truth more straightforward. 

Future work into using numbers of labeled nuclei less than $10,000$ would be of interest but would require a substantial change to the overall process.

\subsection{Position Error Experiments}

Another type of ``weakening'' we could apply to these point annotations is introducing random translation of the sampled points. This would be emulating a particularly lazy or rushed annotator, or an error in the labeling software. This would result in the points being place near but not exactly where the annotator intended. We emulate this effect by adding gaussian noise to the coordinates of the point annotations to introduce a position error of up to $5$ pixels. We expect that this will worsen performance as we will sometimes be labeling background incorrectly but a small position error should not have a drastic effect on our approach.

\begin{table}[]
\begin{tabular}{lll}
Error (Pixels) & Mean Peak Dice & Mean AUROC \\\hline\hline
0                    & 0.71           & 0.944      \\
5                    & 0.635          & 0.926     
\end{tabular}
\caption{Results of adding pixel error to point annotations on Peak Dice and AUROC.}\label{tab:poserror}
\end{table}

\begin{figure*}[!h]
    \centering
    \includegraphics[scale=0.6]{images/Appendix_B/R=3/uncertainty_1516.png.png}
    \includegraphics[scale=0.6]{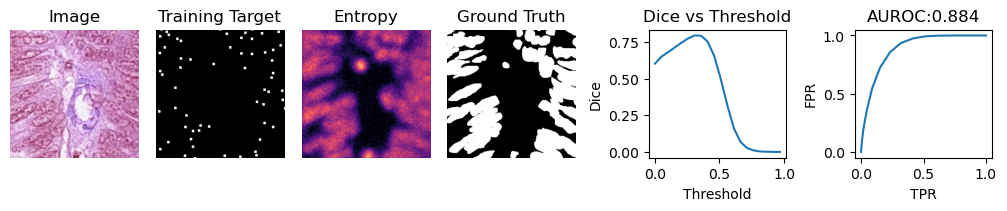}
    \caption{The effect of added noise in point annotation location on entropy output. First row shows no location noise and second row shows error of up to $5$ pixels.}
    \label{fig:AppBexpjitter}
\end{figure*}

Visual outputs of this experiment are shown in \cref{fig:AppBexpjitter} and numerical results are shown in \cref{tab:poserror}. We can see that introducing this error does worsen performance but that the entropy still represents the nuclei distribution fairly well even with this error. We chose to use no error of this type in our experiments as it reduces performance and once again makes comparison with the ground truth much less straightforward. This error type is also not that relevant in many cases since the noise will map onto another point within the nucleus making it less interesting to investigate.

